\documentclass[letterpaper]{article}
\usepackage[preprint]{aaai2027}
\usepackage[hyphens]{url}
\usepackage{graphicx}
\usepackage{natbib}
\usepackage{caption}
\usepackage{booktabs}
\usepackage{enumitem}
\usepackage{amsmath}
\usepackage{amssymb}
\title{Think Shallow, Solve Deep:\\Controlling Recurrent Dynamics for Reliable Test-Time Depth}
\author{
  Ivan Viakhirev\textsuperscript{\rm 1,2},
  Kirill Borodin\textsuperscript{\rm 3,4,5},
  Amirah Almutairi\textsuperscript{\rm 6},
  Serguei Barannikov\textsuperscript{\rm 7},\\
  Maxim Abramov\textsuperscript{\rm 1},
  Grach Mkrtchian\textsuperscript{\rm 3,5}
}
\affiliations{
  \textsuperscript{\rm 1}FRC RAS, Saint-Petersburg, Russia\quad
  \textsuperscript{\rm 2}ITMO, Saint-Petersburg, Russia\\
  \textsuperscript{\rm 3}lab260, Moscow, Russia\quad
  \textsuperscript{\rm 4}BitmanagerAI, Dubai, UAE\quad
  \textsuperscript{\rm 5}MTUCI, Moscow, Russia\\
  \textsuperscript{\rm 6}KAU, Jeddah, Saudi Arabia\quad
  \textsuperscript{\rm 7}CNRS, IMJ-PRG, Paris, France
}

\begin{document}
\maketitle

\begin{abstract}
Recurrent-depth reasoners aim to solve harder problems by iterating their update longer at
test time, but additional iterations can improve, preserve, or degrade an answer. We show
that a measurable property of the trained operator, its finite-time dynamical regime
(estimated as settling, marginal, or drifting), indicates which of these occurs. We give a
sufficient condition for depth-safety: once an operator's per-step displacement is small
relative to the decoder margin, the decoded answer cannot change under further iterations.
Empirically, on algorithmic tasks trained from $800$ unaugmented examples per difficulty
tier, settling operators do not degrade with added depth, and on some tasks convert it
into higher accuracy on harder unseen instances (Sudoku, $0.19$ to $0.34$ past the
training horizon). A single terminal fixed-point objective moves the regime and the depth
behavior together: removing it induces drift and removes the gains, and adding it to a
generic recurrence yields depth-safe extrapolation on carry propagation. We give four
operational criteria for useful test-time depth, use them to catalogue failure modes, and,
as a consistency check, apply the same measurements to Huginn-3.5B, which falls in the
non-settling family.
\end{abstract}

\section{Introduction}

Recurrent-depth reasoners, networks that apply the same block repeatedly to
``think longer'' on harder inputs, are an increasingly popular answer to how small
models might reason. Yet the Tiny Recursive Model \citep{jolicoeur2025trm}
solves Sudoku only after ${\sim}1000$-fold augmentation (without it, the same recurrence
scores $0.00$ under our protocol); HRM \citep{wang2025hrm}
depends on augmentation similarly; and at 3.5B parameters, Huginn's extra recurrence
yields only marginal gains that quickly plateau
\citep{geiping2025huginn,lu2025latentcot}. In all three the architecture has a loop, but training does not make it behave like
an algorithm: added iterations do not reliably improve accuracy on harder instances.

A recurrent reasoner is an iterated map $z_{t+1}=f(z_t,x)$, and whether iterating longer
helps is a property of that map's finite-time dynamics: whether the state settles toward an
input-dependent fixed point, stays marginal, or diverges. A rich theory of
such maps exists (Sec.~\ref{sec:related}), yet reasoning architectures are designed,
trained, and scaled without measuring any of it, compensating with augmentation and
parameters. We make that dynamical regime the object of
study: we measure it on every trained operator, ask what it predicts about test-time
depth, and intervene on it directly.

Our experiments cover a leave-one-out ablation of a controlled-recurrence recipe
(\textsc{CoRe}) and seven re-implemented architectures, all trained under one protocol:
fixed, unaugmented data, $800$ examples per tier, three seeds (three headline
arms re-seeded to $n{\approx}20$). The difficulty
ladders come in two kinds (Sec.~\ref{sec:axes}): ladders that enlarge the input, where
success turns on size-invariance rather than dynamics, and fixed-size ladders
that only deepen the computation.

On the latter the regime predicts the effect of additional iterations: settling operators convert it (Sudoku $0.19\!\to\!0.34$) or hold their
never-trained plateau ($0.92$ on reachability) while marginal and drifting ones stay
flat, collapse, or decay; which architecture settles is task-dependent (on carry it is a
TRM-style attention recurrence, not \textsc{CoRe}); and one terminal objective term
switches the regime in both directions (Fig.~\ref{fig:curves}, Fig.~\ref{fig:causal}).

Our contributions:
\begin{enumerate}
\item \textbf{An operational account of useful test-time depth}: four criteria with
  measurable failure cases, and a proposition fixing what measured settling guarantees
  (depth-safety, and a bound on adaptive-stabilization time) and what stays empirical
  (Sec.~\ref{sec:core}). A measurement suite logs spectral, dynamical, information-flow, and
  topological signals for every run.
\item \textbf{The dynamical regime predicts the fate of test-time depth}:
  settling operators are depth-safe and, on some tasks, convert depth, while marginal and drifting
  operators fail one or both; across our campaign each failure falls into one of four
  measurable modes, and the spectral and topological signatures agree run-by-run
  (Secs.~\ref{sec:core}, \ref{sec:theory}).
\item \textbf{One training term changes both the regime and depth behavior}: the
  terminal fixed-point objective, removed, turns settling into drift and removes conversion
  or safety with the architecture unchanged; added to an alternative recurrence, it
  produces both where added depth had previously degraded the model
  (Sec.~\ref{sec:depth-safety}). We apply the same instruments to Huginn-3.5B
  (Sec.~\ref{sec:huginn}), where they place it in the non-settling family, and a
  label-free objective derived from the same taxonomy repairs its depth-safety
  (digit copying $0.00{\to}1.00$) without installing conversion.
\end{enumerate}

\section{Related Work}
\label{sec:related}

\paragraph{Recurrent-depth reasoners.}
TRM \citep{jolicoeur2025trm} and HRM \citep{wang2025hrm} solve hard puzzles with tiny
recurrent cores but rely on ${\sim}1000\times$ augmentation; Huginn-3.5B
\citep{geiping2025huginn} scales latent recurrence to 3.5B. Input recall
\citep{bansal2022endtoend}, easy-to-hard iteration \citep{schwarzschild2021can}, and
constraint-graph message passing \citep{palm2018rrn} supply recipe components
\citep[cf.][]{esteves2024neuralsolver,knutson2024maze}. Equilibrium Reasoners
\citep{eqr2026} hypothesize reasoning arises from learned attractors; we measure who
learns one and flip it causally. \citet{loopthink2026} see grokking-like transitions
behaviorally \citep[cf.][]{thinkingdeeper2026,latentsurvey2025}, and
\citet{lu2025latentcot} little latent chain-of-thought in Huginn; we supply the
instruments that see both on the inference map. Closest, \citet{labovich2026looped}
analyzes fixed-point stability of looped transformers theoretically; we measure the
trained loop's dynamics per run, across architectures, and switch them causally.

\paragraph{Dynamics of iterated maps in deep learning.}
Dynamical isometry \citep{saxe2014exact,pennington2017resurrecting}, deep equilibrium
models \citep{bai2019deq}, and non-normal transient dynamics in RNNs
\citep{kerg2019nnrnn,orhan2020improved,tarnowski2020transient} supply the map-level
theory. \citet{anil2022path}
showed that path independence correlates with upward generalization in equilibrium
models and requires weight tying and input injection, corroborating our
recall-ablation collapse; our R4 (Sec.~\ref{sec:def}) is its single-initialization face.
Our delta: a failure taxonomy across the full ablation$\times$architecture matrix, a
training-objective switch with two distinguishable consequences (conversion vs.\
depth-safety), and the spectral-anisotropy separation of settled-into-an-algorithm from
merely-settled.

Adaptive halting has its own lineage \citep{graves2016act,banino2021pondernet}; our R2 differs in
demanding that adaptivity convert to correctness (Sec.~\ref{sec:def}).
Edge-of-stability \citep{cohen2021edge} concerns the training-loss Hessian, not the
inference map. A line of work imposes a dynamical property a priori: \citet{bai2021stabilizing}
stabilize DEQ equilibria by regularizing the update's Jacobian, and recent reasoners
follow, including enforced convergence to stable equilibria \citep{cmm2026}, spectral-radius
regularization \citep{stars2026}, and stability-constrained adaptive depth \citep{fprm2026}.
We instead measure which properties are learned, show that
$\sigma_{\max}$-control specifically fails (Sec.~\ref{sec:theory}), and explain
$\rho$-control's success from the non-normal structure of trained operators.

\paragraph{Delayed generalization and progress measures.}
On small algorithmic datasets, generalization can arrive long after the training loss
saturates \citep{power2022grokking}, the generalizing circuit forming gradually,
trackable by hidden progress measures before behavior
\citep{nanda2023progress,barak2022hidden}, in competition with a cheaper memorizing
circuit \citep{varma2023circuit}. Our algorithm-versus-coverage question transfers this
competition to the inference map; Sec.~\ref{sec:corpus} reports which
quantities behave as progress measures for recurrent reasoners.

Finally, persistent homology and topology divergence \citep{barannikov1994framed,barannikov2022rtd} ask whether latent trajectories loop, drift, or settle; we test
three of its hypotheses at campaign scale (Sec.~\ref{sec:theory}).

\section{Setup and Methodology}
\label{sec:setup}


\subsection{Problem statement and the axis we isolate}
\label{sec:axes}
A weight-tied reasoner is an iterated map $z_{t+1}=f_\theta(z_t,x)$ trained for
$T_{\mathrm{train}}$ iterations on a fixed distribution of easier instances. We ask: when does increasing the test-time iteration budget $h_c > T_{\mathrm{train}}$
(i) preserve acquired competence and (ii) convert into accuracy on
harder, out-of-distribution instances?

\begin{figure}[t]
\centering
\includegraphics[width=0.94\columnwidth]{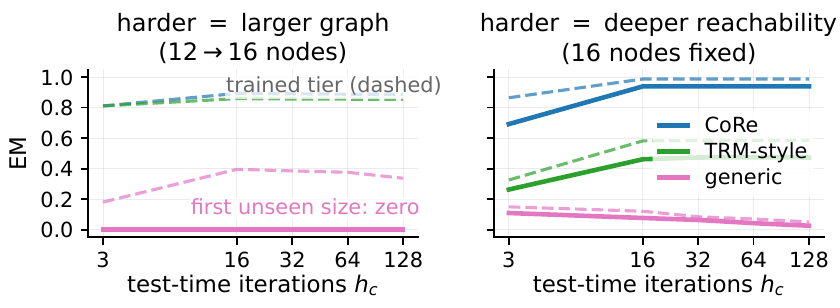}
\caption{Two notions of ``harder,'' one task family. \emph{Left}: larger graphs,
competent on trained sizes (dashed), transfer nothing. \emph{Right}: same-size,
deeper instances, where settling models generalize and plateau.}
\label{fig:scope}
\end{figure}

Two readings of ``harder'' must be separated; the question above is well-posed under
only one. On a fixed-input-size ladder only the number of rounds a
reference algorithm needs grows, so extra iterations are exactly the demanded resource;
on a size/length ladder success is instead governed by the operator's invariance to
input size, a property orthogonal to its dynamics. Figure~\ref{fig:scope} makes the
distinction concrete on one task family. All extrapolation claims therefore live on
fixed-input-size ladders; size/length generalization appears only as scope.

\subsection{Tasks, difficulty ladders, fixed data}
\label{sec:tasks}
Three fixed-input-size ladders with deterministic oracle-round counts carry the primary
experiments:
\begin{itemize}[leftmargin=1.2em,itemsep=0pt,topsep=1pt,parsep=0pt]
\item \textbf{Sudoku}: $9{\times}9$; hole fraction $0.2\!\to\!0.7$, training tiers $0.2/0.35/0.5$;
\item \textbf{reachability}: $16$ nodes; exact BFS depth $2\!\to\!12$, reachable fraction constant by construction;
\item \textbf{addition}: $2{\times}32$ bits, interleaved; longest carry run $2\!\to\!28$.
\end{itemize}
Maze/sorting ladders probe the boundary; size/length ladders support Sec.~\ref{sec:axes};
ARC-1/2 (transductive, no augmentation \citep{chollet2019arc}) is supporting only. Training uses \emph{easy tiers only} (rounds well below
$T_{\mathrm{train}}$), $800$ unique unaugmented examples per tier; evaluation covers the
full ladder ($256$ fixed per tier); splits are materialized once, shared
byte-for-byte; three seeds per configuration.

One run $=$ one trained checkpoint per task$\times$configuration$\times$seed
(evaluations over tiers and depths are repeated observations; parity is excluded as
chance-level for every arm). The headline aggregate is exact-match accuracy (EM; the
fraction of complete outputs satisfying the correctness rule below) on the \emph{first
unseen difficulty tier}, at the training depth and at the best test-time depth
(Table~\ref{tab:main}); means over all unseen tiers are lower and never mixed with it.

\subsection{Metrics: validity oracle and exact match}
Sudoku puzzles with many holes admit multiple valid solutions, so exact match against
the generator's reference solution is ill-posed. We score a prediction correct if and only if it
is a \emph{valid} completion of the input (validity oracle), alongside reference match
$\mathrm{em}_{\mathrm{ref}}$. A solution counter quantifies the ill-posedness: the
uniquely-solvable fraction per tier is $1.00/0.91/0.50/0.06/0.00$; on unseen-hard
Sudoku the valid/reference gap is $5.5\times$ ($0.34$ vs.\ $0.062$; Sec.~\ref{sec:core}
confronts the surplus). Aggregates are per-puzzle exact match (never per-cell, which
saturates misleadingly) and \emph{solve steps} (the first iteration at which the decoded answer stops
changing); on reachability and addition the oracle-round count is deterministic from
the input, so adaptivity is checked against required rounds directly, free of Sudoku's
caveat.

\subsection{Protocol, test-time depth, architectures}
\label{sec:arch-protocol}
All ladder models train at $16$ recurrent steps with deep supervision ($4$ supervised
unrolls), AdamW, gradient clipping $0.5$, batch $64$, $4$--$8$k steps (task-dependent),
$d{=}256$; at evaluation we sweep test-time depth $h_c\in\{3,16,32,64,128\}$, the
``think longer'' axis (per-task settings in the appendix). We compare
\emph{mechanisms}, not checkpoints. Seven architectures (TRM-style
\citep{jolicoeur2025trm}, UT/URM \citep{dehghani2019universal}, FPRM \citep{fprm2026},
EqR, DEQ \citep{bai2019deq}, Neural GPU \citep{kaiser2016neuralgpu}, SE-RRM) are
re-implemented at matched size and budget and trained on the identical data files,
protocol, seeds, and logging as every ablation arm (not authors' checkpoints). The one
pretrained model, Huginn-3.5B, is measured via forward hooks (measure-only;
Sec.~\ref{sec:huginn}).

\subsection{The measurement toolkit}
\label{sec:toolkit}
Every run logs, at the trained model on held-out inputs, five families of signal:
\emph{spectral} (step-Jacobian $\sigma_{\max}$, $\rho$, Henrici non-normality
\citep{henrici1962bounds}, random-probe gain); \emph{dynamical} (top-8 finite-time
Lyapunov spectrum via Benettin QR \citep{benettin1980lyapunov},
$\sigma_{\mathrm{eff}}{=}\exp(\lambda_{\max})$, Kaplan--Yorke dimension, fixed-point
residual, settle ratio); \emph{information-flow} (logit-lens rank of the correct answer
\citep{belrose2023tunedlens}); \emph{topological} (Ripser
\citep{bauer2021ripser}: Betti bars, winding, effective dimension); and
\emph{training-time} ranks (full inventory and estimator scope in the appendix).
Regime labels: \textbf{settle} ($\lambda_{\max}{<}-0.05$), \textbf{marginal}
($|\lambda_{\max}|\le0.05$), \textbf{drift} ($\lambda_{\max}{>}0.05$).

\section{The Recipe, Mechanistically}
\label{sec:recipe}

Each recipe component removes one measured way for the loop to avoid becoming an
algorithm; we present each through its ablation failure (Sec.~\ref{sec:core}).

\paragraph{Input recall: keep the problem in the loop.}
All models condition on the embedded input once per step; the recipe \emph{re-injects}
it before each update \citep{bansal2022endtoend}. Removing
only this extra injection collapses the trained operator to an \emph{identity}
($\sigma_{\max}=1.0000$ to four decimals, EM $0.00$): the loop discovers an input-insensitive identity solution; the reinforced injection makes the fixed point strongly
\emph{input-dependent}, the minimal requirement for the limit to encode an answer.

\paragraph{Structured message passing: the computation graph is the constraint graph.}
Our step mixes states along rows, columns, and boxes (task-appropriate neighborhoods
elsewhere) \citep{palm2018rrn}. Removing it keeps the model able
to settle (the no-structure arm contracts hardest of all) and it still scores
$0.00$: settling is necessary, not sufficient. Structure buys what the loop
computes; dynamics control buys that it converges.

\paragraph{Fixed-point objective: make the answer a fixed point.}
We penalize the displacement of the final update at the supervised horizon,
$\lVert z_{T}-z_{T-1}\rVert=a_{T}\,\lVert f(z_{T-1},x)-z_{T-1}\rVert$, i.e.\ the
fixed-point residual evaluated at the penultimate state, scaled by the learned update
gate. This teaches the map to make its answer a fixed point (the gate does not trivially
satisfy the penalty by closing; Appendix~\ref{supp:gate}). This is the principal objective term evaluated by
ablation: removing only this term leaves the architecture, data, and
$\sigma_{\max}\,({\approx}5.6)$ intact but flips
the trained regime from settle to drift ($\lambda_{\max}\,{-}0.06\!\to\!{+}0.10$,
Kaplan--Yorke dimension $0\!\to\!8$), and extrapolation dies ($0.34\!\to\!0.03$ on unseen-hard Sudoku): the attractor is what
makes ``think longer'' safe: iterations beyond the horizon approach the answer
instead of leaving it (Sec.~\ref{sec:core}).

\paragraph{Remaining stabilizers and their measured scope.}
The remaining components survive ablation with their means intact and are claimed
only for what they measurably buy (Appendix~\ref{supp:fulltable}, Table~\ref{tab:full}). Orthogonal
initialization \citep{saxe2014exact} leaves accuracy unchanged ($0.34$ vs.\ $0.34$) but
selects a different solution family (weight stable rank ${\approx}70$--$89$ vs.\
$20$--$30$); the adaptive update magnitude and gradient clipping buy seed stability
(extrapolation s.d.\ $0.001$ vs.\ $0.023$; $\pm0.01$ vs.\ $\pm0.13$) at task-dependent
cost (reachability trainability $0.99$ vs.\ $0.69$); deep supervision the same. One
popular component fails scrutiny: a lower learning rate is not essential: the
higher-LR ablation outperforms the recipe ($0.42$ vs.\ $0.34$), so we report LR
as ``matched to scale'' and keep the pre-registered headline. The recipe's effect comes
from recall, structure, and the fixed-point objective.

\section{Why Some Networks Learn Recursion, and Some Do Not}
\label{sec:core}

\subsection{What we mean by learned recursion}
\label{sec:def}
We use four operational criteria for whether a trained model makes useful and safe use
of test-time depth on a task family. The criteria are behavioral and do not by themselves
establish that a particular algorithm was learned; a model meets them if all four hold:

\begin{itemize}
\item[\textbf{R1}] \emph{Extrapolation by iteration.} On instances harder than anything
  seen in training, accuracy \emph{increases} with test-time iterations beyond the training
  depth. (Ruled out: distributional coverage, whose depth curve is flat.)
\item[\textbf{R2}] \emph{Difficulty-adaptive stabilization that converts}. The
  suffix-stabilization step (the earliest iteration after which the decoded answer
  never changes through the evaluation horizon) tracks instance difficulty \emph{and}
  the stabilized answer is correct. (This is a retrospective diagnostic, not an online
  halting rule. Counterfeit: any non-degenerate map's latents move longer on harder
  inputs, so adaptivity alone proves nothing.)
\item[\textbf{R3}] \emph{Incremental computation.} The correct answer forms gradually
  across iterations, visible as a monotone descent of its logit-lens rank. (Counterfeit:
  guess-then-polish, with the answer present at step 1.)
\item[\textbf{R4}] \emph{Convergent answer.} The \emph{decoded answer} converges:
  further iterations neither change nor degrade it. The \emph{strong form} is a latent
  fixed point ($\sigma_{\mathrm{eff}}{<}1$, residual $\to0$); the \emph{weak form} (answer
  frozen while the latent moves in directions the readout ignores) is realized by carry-transport
  solutions (Sec.~\ref{sec:depth-safety}); cf.\ path independence \citep{anil2022path}.
  (Ruled out: depth-fragility and guess-freezing, the latter formalized by
  Prop.~1(b), both excluded by conjunction with R1--R3.)
\end{itemize}

The criteria are deliberately behavioral (none mentions a spectrum or an
exponent), while the dynamical quantities of Sec.~\ref{sec:toolkit} are measured
independently of them; that the two coincide run-by-run is this paper's empirical claim.
A model with high accuracy that fails R1--R4 uses recurrence without measurable benefit
from added depth: the extra iterations do not contribute, and the solution is typically
obtained by coverage \citep[cf.][]{varma2023circuit}.

One relation among the four is not empirical but forced, and it fixes what ``measured
settling'' can and cannot buy.

\medskip\noindent\textbf{Proposition 1 (stabilization dichotomy).}
\emph{Let $\delta_t=\lVert z_{t+1}-z_t\rVert$ be the per-step displacement (the quantity
the settle ratio measures), let the decision margin $\mu(z_t)$ be the largest $r$ with the
decoder $g$ constant on the open ball $B(z_t,r)$, and let $R_t=\sum_{u\ge t}\delta_u$ be
the remaining path length. \textbf{(a)} If $R_t<\mu(z_t)$, then $g(z_s)=g(z_t)$ for all
$s\ge t$. \textbf{(b)} If $R_t=\infty$, no finite window of answer-constancy certifies the
next step; this can occur even when $\delta_u\to0$.}

\noindent\emph{Proof.} \textbf{(a)} For any $s\ge t$, the triangle inequality gives
\begin{equation}
\lVert z_s-z_t\rVert\ \le\ \sum_{u=t}^{s-1}\delta_u\ \le\ R_t\ <\ \mu(z_t),
\label{eq:freeze}
\end{equation}
so $z_s$ lies in the margin ball of $z_t$ and $g(z_s)=g(z_t)$, uniformly in $s$.
\textbf{(b)} The map $f(z)=z-(c,0,\dots)$ with $g=\mathrm{sign}(z^{(1)})$ and generic
$c,D$ has $\delta_t=c>0$, so $R_t=\infty$, yet the answer holds for $\lfloor D/c\rfloor$
steps and then flips, with $D$ arbitrary; a divergent sum with $\delta_u\to0$
(e.g.\ $\delta_u=1/u$) behaves likewise. $\square$

\noindent The premise of (a) is measured, not assumed: $R_t$ is the tail of the logged
displacement, and geometric decay gives an explicit freeze time growing with the
transient scale and with $1/\mu$ (Appendix~\ref{supp:proofs}). Branch (a) makes R4 depth-safety and the
adaptive-timing half of R2 consequences of \emph{measured} settling (while decay
persists), not of correctness, so settle-but-wrong remains possible, and both
R1 and R2's conversion requirement stay empirical. Branch (b) is the formal definition of guess-freezing,
whose premise Huginn meets (Sec.~\ref{sec:huginn}). The rest of Secs.~\ref{sec:core}--\ref{sec:theory} measures the premises and consequences of Proposition~1 on small
controlled models, and Sec.~\ref{sec:huginn} shows branch (b) at scale.

\subsection{Test-time depth across ladders and architectures}
\label{sec:ttscale}
\begin{figure*}[t]
\centering
\includegraphics[width=0.78\textwidth]{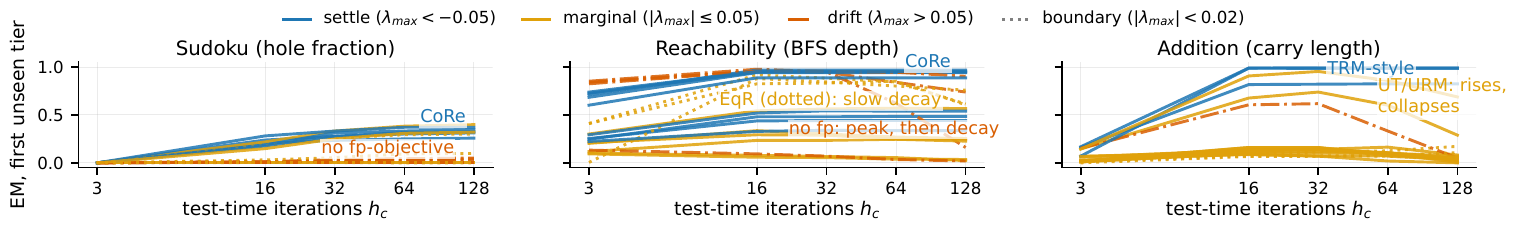}
\caption{Depth curves behind Table~\ref{tab:main}: EM on the first unseen tier; one
line $=$ one run, colored by \emph{measured} regime (boundary dotted, drift
dash-dotted): settling runs rise and plateau, marginal rise and collapse or stay flat,
drifting decay here or on deeper tiers (Sec.~\ref{sec:depth-safety}); which
architecture settles differs per ladder.}
\label{fig:curves}
\end{figure*}

Across the three primary ladders, three regularities emerge in the depth curves
(Table~\ref{tab:main}, Figure~\ref{fig:curves}; each run colored only by measured
regime, boundary dotted, the appendix aggregates by regime).

\textbf{Learnability is not the bottleneck}: most arms fit training
($\mathrm{EM}\ge0.97$ in $32/51$ cells of Table~\ref{tab:full}); rows below the competence gate
($\mathrm{EM}<0.3$) are excluded from extrapolation verdicts.

\textbf{Conversion tracks the regime, not the architecture}: on Sudoku the settling recipe family reaches
$0.34{\pm}0.04$ (variants up to $0.42$; pre-registered headline kept); on reachability
(whose ladder tops out below the horizon, so beyond-horizon conversion is not in play)
the settling family generalizes to never-trained depths ($0.92{\pm}0.10$) on a flat
plateau; on carry the settling run is TRM-style ($0.82{\pm}0.16$: its attention step
realizes the carry algorithm our local mixing cannot, its halting time nearly matching
the oracle's carry length), while
marginal runs rise then collapse (UT/URM $0.95$@$32\to0.05$@$128$). TRM also shows the
link within its Sudoku seeds: the settling seeds extrapolate while near-marginal
ones do not.

Augmentation cuts both ways: applied to
our TRM re-implementation (its native data regime, $2\times$ budget) it lifts Sudoku to
$0.16$ and recovers a previously unsuccessful seed, an optimization aid (no claims about TRM's ceiling);
applied to the recipe it leaves training tiers and the settling regime intact but
collapses extrapolation ($0.34\!\to\!0.05$) while flattening the Lyapunov spectrum
(Sec.~\ref{sec:corpus}): coverage neither needs nor builds the anisotropic operator
that extrapolation runs on.

\textbf{Each ablation damages the property it targets}: the fixed-point objective is
analyzed causally in Sec.~\ref{sec:depth-safety}; removing structure or recall removes
trainability itself (reachability $0.99\!\to\!0.15$). Across the 304 original-ladder
runs, \emph{all 19 extrapolating constraint-propagation runs are non-drift}
($\lambda_{\max}$ median $-0.047$ vs.\ $+0.009$, $p{=}7{\times}10^{-6}$).

\begin{table*}[t]\centering
\setlength{\tabcolsep}{4.2pt}
\resizebox{\textwidth}{!}{%
\begin{tabular}{l ccc ccc ccc}
\toprule
 & \multicolumn{3}{c}{Sudoku (hole fraction)} & \multicolumn{3}{c}{Reachability (path length)} & \multicolumn{3}{c}{Addition (carry length)}\\
\cmidrule(lr){2-4}\cmidrule(lr){5-7}\cmidrule(lr){8-10}
 & learn\,$\uparrow$ & extrapolate\,$\uparrow$ & overthink\,$\downarrow$ & learn\,$\uparrow$ & extrapolate\,$\uparrow$ & overthink\,$\downarrow$ & learn\,$\uparrow$ & extrapolate\,$\uparrow$ & overthink\,$\downarrow$\\
\midrule
\textsc{CoRe} (full recipe) & \textbf{1.00} & 0.19$\to$\,\textbf{0.34}\,{\scriptsize$\pm$0.04} & \textbf{0} & 0.98 & 0.92$\to$\,0.92\,{\scriptsize$\pm$0.10} & \textbf{0} & 0.99 & 0.13$\to$\,0.17\,{\scriptsize$\pm$0.15} & 0.05\\
\addlinespace[2.5pt]
~~$-$\,structure & 0.00 & --- & n/a & 0.15\,{\scriptsize$\pm$0.02} & --- & n/a & \multicolumn{3}{c}{\textit{= CoRe (no spatial structure on 1-D)}}\\
~~$-$\,extra input injection & 0.29\,{\scriptsize$\pm$0.01} & --- & n/a & 0.23 & --- & n/a & 0.04\,{\scriptsize$\pm$0.02} & --- & n/a\\
~~$-$\,fixed-point objective & \textbf{1.00} & 0.01$\to$\,0.03\,{\scriptsize$\pm$0.01} & 0.04 & \textbf{1.00} & 0.97$\to$\,\textbf{0.97}\,{\scriptsize$\pm$0.02} & 0.37$^\dagger$ & \textbf{1.00} & 0.14$\to$\,0.15\,{\scriptsize$\pm$0.02} & 0.81$^\dagger$\\
\addlinespace[2.5pt]
generic recurrence & 0.00 & --- & n/a & 0.17\,{\scriptsize$\pm$0.01} & --- & n/a & \textbf{1.00} & 0.11$\to$\,0.12\,{\scriptsize$\pm$0.02} & 0.84$^\dagger$\\
TRM-style & 0.94\,{\scriptsize$\pm$0.18} & 0.14$\to$\,0.21\,{\scriptsize$\pm$0.11} & \textbf{0} & 0.70\,{\scriptsize$\pm$0.17} & 0.55$\to$\,0.56\,{\scriptsize$\pm$0.22} & 0.01 & \textbf{1.00} & 0.80$\to$\,\textbf{0.82}\,{\scriptsize$\pm$0.15} & \textbf{0.02}\\
UT/URM & 0.97\,{\scriptsize$\pm$0.06} & 0.00$\to$\,0.00 & 0.18 & 0.43\,{\scriptsize$\pm$0.09} & 0.28$\to$\,0.28\,{\scriptsize$\pm$0.07} & 0.09 & 0.99\,{\scriptsize$\pm$0.02} & 0.66$\to$\,0.72\,{\scriptsize$\pm$0.21} & 0.59$^\dagger$\\
FPRM & 0.53\,{\scriptsize$\pm$0.40} & 0.00$\to$\,0.01\,{\scriptsize$\pm$0.01} & 0.04 & 0.31\,{\scriptsize$\pm$0.05} & 0.20$\to$\,0.20\,{\scriptsize$\pm$0.03} & 0.11 & 0.82\,{\scriptsize$\pm$0.26} & 0.27$\to$\,0.28\,{\scriptsize$\pm$0.25} & 0.80$^\dagger$\\
EqR & \textbf{1.00} & 0.01$\to$\,0.01\,{\scriptsize$\pm$0.01} & 0.13 & 0.99\,{\scriptsize$\pm$0.01} & 0.88$\to$\,0.89\,{\scriptsize$\pm$0.03} & 0.24$^\dagger$ & \textbf{1.00} & 0.12$\to$\,0.12\,{\scriptsize$\pm$0.03} & 0.72$^\dagger$\\
DEQ & \textbf{1.00} & 0.00$\to$\,0.00 & \textbf{0} & 0.26\,{\scriptsize$\pm$0.04} & --- & n/a & 0.99\,{\scriptsize$\pm$0.01} & 0.19$\to$\,0.47\,{\scriptsize$\pm$0.19} & 0.07\\
Neural GPU & 0.00 & --- & n/a & 0.76\,{\scriptsize$\pm$0.08} & 0.54$\to$\,0.54\,{\scriptsize$\pm$0.11} & 0.68$^\dagger$ & --- & --- & ---\\
\bottomrule
\end{tabular}}
\caption{Three questions, kept separate (mean$\pm$sd): the primary
ablations and all architectures (Appendix~\ref{supp:fulltable}, Table~\ref{tab:full} for six further ablations,
none switching mode). \textbf{learn}\,$\uparrow$: in-distribution EM at best depth.
\textbf{extrapolate}\,$\uparrow$: first-unseen-tier EM at training depth $\to$ best
over $h_c{\le}128$, selected per run on the evaluation set (R1 conversion). \textbf{Bold} $=$ best value
per column, ties included (max for learn/extrapolate, min for overthink; gated cells
excluded).
\textbf{overthink}\,$\downarrow$: worst over tiers of best EM
$-$ EM@128 ($\dagger$ $=$ severe ${\geq}0.2$ loss, undesirable). EM${<}0.3$ fails the competence
gate: extrapolate/overthink are n/a (nothing to convert or lose). Reachability's rounds
stay below the training depth, so its middle column is generalization$+$safety, not
conversion; ``---'' marks undefined or gated cells. Naming per Sec.~\ref{sec:recipe}. \textsc{CoRe}, $-$fixed-point objective, and TRM-style cells carry $n{\approx}20$ seeds, architecture rows $n{=}10$, the remaining ablation rows $n{=}3$.}
\label{tab:main}\end{table*}

\begin{figure*}[t]
\centering
\includegraphics[width=0.78\textwidth]{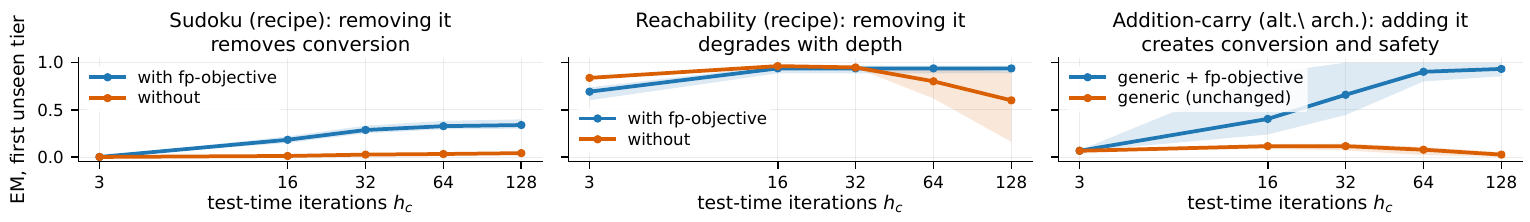}
\caption{One training term, three consequences (mean and seed range, first unseen tier).
Removing the fixed-point objective destroys conversion (Sudoku, left) or safety
(reachability, middle); adding only this term to the unchanged generic recurrence
creates both on addition-carry (right): $0.12\!\to\!0.88$, damage $0.91\!\to\!0$.}
\label{fig:causal}
\end{figure*}

\subsection{One task, every way to fail it}
\label{sec:onetask}
The patterns of Table~\ref{tab:main} become mechanically transparent on Sudoku, where
the target algorithm (constraint propagation) is known. The full recipe passes all four criteria: it climbs from $0.00$ (shallow) and $0.19$
(training depth) to $0.34$/$0.11$ on unseen tiers 3/4 (R1); solve steps grow
monotonically ($2.8\!\to\!19.6$), commensurate with oracle rounds where propagation
terminates and monotone on the $94\%$-under-determined hard tiers (R2); rank
decreases monotonically toward zero ($0.33\!\to\!0.05$; R3); and all eight estimated
leading exponents are negative (strong-form R4).

Is a legal completion of an under-determined board still ``harder''? It is partly a
different competence (constraint satisfaction, not unique deduction), so we split
the credit: reference-EM covers essentially every uniquely determined tier-3 puzzle
($0.062$ vs.\ $0.059$ uniquely solvable), and the validity surplus is not free (every
baseline scores $0.00$ under the same oracle), arrives through iteration, and recurs on
the ambiguity-free reachability ladder.

Every other outcome in the campaign is a specific, measurable violation:
\emph{identity} (no-recall: instant convergence to ``do nothing'',
$\sigma_{\max}{=}1.0000$, EM $0.00$); \emph{marginal} (generic, UT/URM, FPRM: an
exactly normal, norm-preserving wander that fits easy tiers, depth curves flat or
declining); \emph{drift} (no fixed-point objective: $\lambda_{\max}{=}{+}0.10$, extra
steps hurt); \emph{settle-but-wrong} (no-structure; TRM on ARC: dynamics right, fixed
point wrong, settling \emph{isotropically}, spread ${\le}0.08$, where true learners
are anisotropic, the instrument of Sec.~\ref{sec:corpus}); full numbers in
Appendix~\ref{supp:fulltable}. The remaining ablation arms degrade gradually rather than switching mode, the
sense in which recall, structure, and the fixed-point objective are the necessary
components.

\subsection{Depth-safety and overthinking: R4 in both forms}
\label{sec:depth-safety}
The clearest intervention is Figure~\ref{fig:causal}: removing the terminal fixed-point
objective, which jointly moves the regime and the depth behavior (i.e.\ whether the
premise of Prop.~1(a) is met), destroys a different half of the thesis per ladder. On
Sudoku it destroys \emph{conversion} (the drifting ablation never rises,
$0.34\!\to\!0.03$); on reachability it leaves the peak ($0.97$) but destroys
safety (the drifting run loses up to $0.37$ EM by $h_c{=}128$ while the settling
recipe holds $0.92\!\to\!0.92$). The intervention also runs \emph{constructively}: adding only
this term to the unchanged generic recurrence turns the architecture carry propagation
destroys (overthink $0.91$) into a depth-safe extrapolator (best $0.88$, $18$ seeds,
damage $0$); UT/URM moves the same way on Sudoku ($0\!\to\!0.16$--$0.19$) and FPRM on
carry, while EqR does not respond (it hard-wires this penalty at a weight whose residual
never resolves; Appendix~\ref{supp:interventions}) and DEQ stops fitting training altogether
($10/10$ seeds), the boundaries of the intervention.

The payoff requires a \emph{latent} attractor, not a frozen output: a control penalizing only the decoded answer's change (latent free)
yields neither conversion nor safety (Sudoku extrapolates to $0.00$, latent never settling;
reachability peaks then decays, worst-tier damage up to $1.00$; Appendix~\ref{supp:interventions}), and a
weight sweep shows depth-safety turning on precisely at the settling crossing, with
conversion growing as the weight rises further; settling is necessary, not
sufficient.

Training deeper
($T_{\mathrm{train}}{=}32$) shows a same-direction gain past its own horizon
($0.43{\pm}0.03$@$32\to0.48{\pm}0.03$@$128$; $n{=}15$ seeds): a longer horizon does not substitute
for test-time iteration (pre-registered $T{=}16$ stays the headline). Without an attractor a
model can reach a correct answer transiently but not keep iterating safely: branch (b)
of Prop.~1, the operational meaning of overthinking.

R4 is about the answer: a \emph{size}-ladder experiment (off the extrapolation axis,
Sec.~\ref{sec:axes}) shows why the latent fixed point is R4's strong, not only, form:
on addition-by-\emph{length} the generic recurrence freezes its answer within
${\sim}5$ steps \emph{while its latent never settles}
($\sigma_{\mathrm{eff}}{=}1.02$, weak-form R4), yet on prefix sums the same
statistics give depth-fragility ($1.00\!\to\!0.65$; Appendix~\ref{supp:weakr4}). So latent spectra
alone cannot separate weak-form R4 from depth-fragility; the difference lives in
answer space, where R4 is stated.

The strong form makes ``think longer'' a safe
default: among all 164 competent runs, not one of the 17 settling runs
loses more than $0.004$ EM to depth, while $30\%$ marginal and $22\%$ drifting lose
more than $0.1$, up to $0.98$ (Neural GPU on mazes, $p{=}10^{-3}$). Two notes: under a
task$\times$configuration cluster bootstrap the safety separation survives (excess
unsafe $0.14$, CI $[0.06,0.24]$) while pooled conversion is not cluster-significant
(conversion rests on the depth curves and the causal switch), and the $\pm0.05$
threshold is not load-bearing (Appendix~\ref{supp:nonnormal}). On tasks where the recipe fails outright
(addition, sort), its own rank curves rise with iterations ($0.74\!\to\!0.94$), the
information-level image of drift; the same fates reappear behaviorally at 7B
(Sec.~\ref{sec:cot}).

In sum: learning recursion is passing R1--R4, and across the corpus the runs that do so
are consistently settling-but-barely and strongly non-normal, with the settling
architecture differing by task.

\section{The Shape of Learned Recursion}
\label{sec:theory}


\subsection{Hypotheses and the non-normality account}
\label{sec:nonnormal}
A topology-first program \citep{barannikov2022rtd} predicts settle/loop/drift classes
(H1), winding tracking depth (H2), and that a purely contracting map cannot count (H3).
Verdicts (Appendix~\ref{supp:hypotheses}): \textbf{H1} settle/drift ubiquitous and causally separable,
loops nowhere; \textbf{H2} unsupported (solvers move radially, 99th-pct
$|$winding$|$ $0.71$ turn); \textbf{H3} supported, with a sharpened mechanism.

Why is $\sigma_{\max}$ a correlate but not a lever (Prop.~2)? Successful operators are
strongly non-normal ($\sigma_{\max}/\rho\approx4$--$5.4$): the top singular value is a
\emph{transient} amplification, not an asymptotic rate ($\rho{<}1$); random-probe gain
stays ${\le}1$, so computation lives in a \emph{thin} amplifying cone inside a globally
contracting map. A typical-gain sweep (\citet{bai2021stabilizing}'s regularizer as a probe) confirms
it: pinning the typical gain at its decaying level is harmless; forcing it to expand
($\tau{\geq}2$) blows $\sigma_{\max}$ to $14$--$33$ and destroys training
(Appendix~\ref{supp:hypotheses}). Regularizing $\rho$ (asymptotics only) is the control
that helps at scale \citep{stars2026}; non-normality alone does not (Neural GPU drifts
to zero at $\sigma_{\max}/\rho{=}17.7$).

\begin{figure*}[t]
\centering
\includegraphics[width=0.72\textwidth]{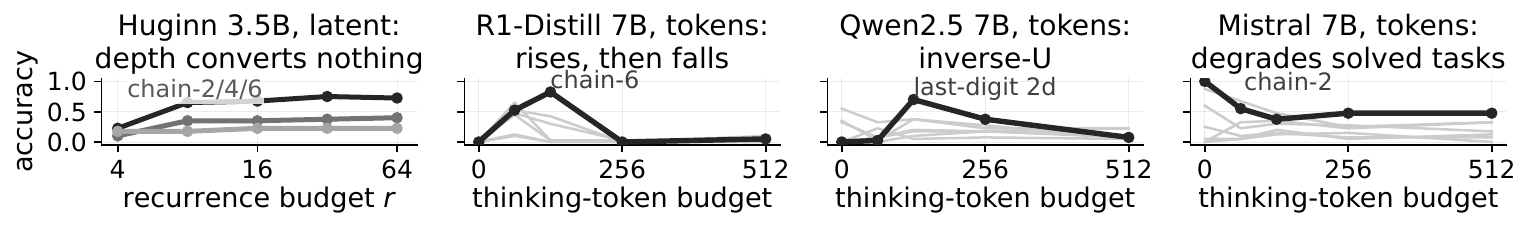}
\caption{Recurrence at scale, one prompt bank. \emph{Left}: a latent depth knob
(Huginn-3.5B) saturates without converting depth. \emph{Right three}: token-budget
knobs on open 7B reasoners (bold $=$ illustrative family, $n{=}40$/point); none shows
the settling fate.}
\label{fig:cot}
\end{figure*}

\subsection{What the full corpus adds}
\label{sec:corpus}
\paragraph{Spectral anisotropy: the signature that separates algorithms from mere
convergence.}
The shape of the Lyapunov spectrum is the strongest single predictor we measured: algorithm-%
learners develop a structured spectrum, a hierarchy of contraction rates
(\textsc{CoRe} on Sudoku: spread $\lambda_1{-}\lambda_8{=}0.41$), while every
failure mode is spectrally degenerate: eight nearly identical exponents, whether
pinned at zero (identity), at ${+}0.02$ (marginal), or at $-0.17$ (no-structure:
settled hard, isotropically); Appendix~\ref{supp:spectra} overlays the full landscape.

The spread alone predicts extrapolation with AUC $0.846$ across 304 runs, the best
single dynamical quantity we measured though on par with training accuracy itself
($0.851$): a corpus-level correlate whose value is the within-settle separation, not a
universal discriminator (reachability's extrapolators settle at spreads $0.07$--$0.14$).

The Kaplan--Yorke dimension is bimodal ($47\%$ exactly $0$, $48\%$ saturated),
giving R4's strong form its practical reading: a learner of recursion has all eight
estimated leading exponents negative. Within the settling regime the spread separates
runs that settled into an algorithm from runs that merely settled (AUC $0.872$,
$p{=}10^{-3}$), a distinction no other instrument could make, including on
TRM-trained ARC models (settled, flat, wrong); the augmentation control of
Sec.~\ref{sec:ttscale} is the same flattening shown causally. We read this as functional differentiation: an iterated algorithm needs directions
with different jobs (one instrument-level exception in Appendix~\ref{supp:corpus}).

\paragraph{Three further regularities (Appendix~\ref{supp:corpus}).}
Live spectra and post-hoc geometry agree run-by-run ($\sigma_{\mathrm{eff}}$ vs.\
settling $r{=}0.75$, $n{=}325$; replicated $\rho{=}0.69$, $n{=}1{,}206$); topology sees
compactness, not loops ($92/24/2\%$ across regimes; per-instance AUC $0.80$--$0.98$,
Remark~1); and the algorithm forms late (rank $33{\to}53$ while siblings stay flat),
the inference-map face of delayed generalization (Sec.~\ref{sec:related}); dashboard
in Appendix~\ref{supp:dashboard}.

\section{Recurrence at Scale}
\label{sec:huginn}

This section tests the taxonomy at scale, then repairs a depth pathology of a
pretrained recurrent LLM with a label-free objective.
Huginn-3.5B \citep{geiping2025huginn} iterates the same block on its latent state; a
chain-of-thought reasoner iterates in token space, where only behavior compares;
neither is a performance baseline.

\subsection{Latent recurrence: Huginn-3.5B}
\textbf{R1}/\textbf{R3} are answered in prior work: recurrence $4\!\to\!32$ yields
marginal gains far below chain-of-thought (Figure~\ref{fig:cot}, left), no
structured latent chain found \citep{lu2025latentcot}.

\textbf{R4}: across per-token latent trajectories on 7 task categories ($2{,}585$
tokens), not one settles ($\sigma_{\mathrm{eff}}{=}1.000{\pm}0.003$; layer
normalization pins only the radial direction, not the read displacement):
\emph{normalized-marginal}. \textbf{R2}: the
answer freezes, but freezing does not track difficulty: guess-freezing (Prop.~1b),
not a computation completing (Appendix~\ref{supp:layernorm}; \citealp{lu2025latentcot}); by R1--R4 the
loop gains nothing from added depth.

The base's depth is unsafe, not just inert: on procedural single-token probes
accuracy peaks shallow and collapses (carry: $0.69$ at $r{=}8$, $0.00$ for
$r{\ge}32$), every input freezing onto the same wrong token. A settling event itself, guess-freezing
evades answer-stability halting; a 30-example depth calibration recovers the peak at
the price of labels and a per-task depth the repair below does not need (Appendix~\ref{supp:repair}).

\paragraph{Repairing depth-safety without labels.}\label{sec:repair}
Guided by the taxonomy, a rank-16 LoRA adapter on the recurrent core is trained
with one label-free latent objective: pull deep states, $r{\sim}U[16,128]$, toward the
frozen base's own shallow state at $r{=}8$ (two synthetic anchor domains, $2{,}000$
steps); the fixed-point term itself does not transfer to this scale, only its
latent-anchoring form (Appendix~\ref{supp:repair}). The patch restores depth-safety, not new
capability, on tasks never
seen in training: digit copying, lost by the base at $r{\ge}4$, reaches
$1.00{\pm}0.00$ at $r{=}128$ with input-dependent answers ($10/10$ seeds); letter
repetition recovers $0.00{\to}0.51{\pm}0.13$ (a carry plateau: stabilization only;
Appendix~\ref{supp:repair}).
Conversion never appears, and the depth-hard pointer-chase task stays unlearnable
under every intervention tried, making the no-benefit-from-depth verdict causal;
transfer follows the anchors' operation family across alphabets, with two failure
laws measurable in advance (Appendix~\ref{supp:repair}).

\subsection{Token recurrence: the same fates}
\label{sec:cot}
Chain-of-thought gives a reasoner its own depth knob; on one prompt bank three open
models reproduce the non-settling fates (Figure~\ref{fig:cot}): the distilled reasoner is destroyed by more of the budget it needs
($0.82\!\to\!0.00$) and the non-reasoning model degraded by thinking it does not
need ($1.0\!\to\!0.47$; a shared failure shape, not mechanism identity;
\citealp{chen2024overthink,su2025underthinking,sui2025stopoverthinking}).
Both substrates overthink; the case for latent depth is the handle: only
the latent loop exposes a state one can measure and repair (Sec.~\ref{sec:repair}).

\section{Limitations}
\label{sec:limitations}
\textbf{Synthetic generators.} Sudoku boards come from one construction's symmetry
orbit, so the headline is extrapolation on that generator; the matching reachability
plateau shows the claim does not rest on Sudoku's scoring. \textbf{Controlled protocol.}
Budget-matched re-implementations estimate no ceiling, which isolates the regime
from tuning confounds; producing a settling operator is not guaranteed
(on mazes, a seed lottery); where obtained, the outcome follows the regime.
\textbf{Small scale.} Models are
tiny and tasks synthetic, yet the same instruments place Huginn-3.5B in the marginal
family; the scale repair (Sec.~\ref{sec:repair}) is operation-family-scoped, one
checkpoint. \textbf{Mediation.} We identify the \emph{objective} as the causal lever and the \emph{latent} attractor
as its needed target (answer-only control, Sec.~\ref{sec:depth-safety}), not the
regime as sole mediator; finite-time estimates bound adjectives, not regime
separations (Appendix~\ref{supp:nonnormal}).

\section{Conclusion}
\label{sec:conclusion}
Learned recursion is a dynamical property: a non-normal operator settling in an
attractor the objective creates and structure aligns with the task; failures fall
into four measurable modes. Depth-safety is provable at small displacement (Prop.~1);
conversion stays empirical and task-dependent. The measurements, and the repair they
prescribe, carry to 3.5B: whether added depth helps is a measurable property of the
dynamics.

\bibliography{paper}

\twocolumn[{\centering\Large\bfseries Supplementary Material\par\vspace{1.5em}}]
\appendix

\section{Propositions and proofs}
\label{supp:proofs}
This section gives the geometric-decay corollary of Proposition~1, the trajectory-%
informativeness remark, and Proposition~2 (all referenced from the main text). Notation
as in the main text: $\delta_t{=}\lVert z_{t+1}{-}z_t\rVert$, decision margin $\mu(z)$,
remaining path $R_t{=}\sum_{u\ge t}\delta_u$; $g$ is the (argmax) decoder.

\paragraph{Corollary 1 (explicit freeze time under geometric decay).}
If $\delta_u\le C\gamma^u$ with $\gamma{<}1$, then $R_t\le C\gamma^t/(1{-}\gamma)$, and by
Proposition~1(a) the decoded answer is frozen for all $t\ge t^\star$, where
\[
t^\star=\max\!\Big(0,\ \Big\lceil \log\tfrac{C}{(1-\gamma)\,\mu}\Big/\log\tfrac1\gamma\Big\rceil\Big),
\]
obtained by solving $C\gamma^t/(1{-}\gamma)<\mu$ (open ball, strict). The freeze time
grows with the transient scale $C$ (longer propagation) and with $1/\mu$ (smaller margin,
harder instance); with $\gamma{=}e^{-|\lambda|}$ the leading order is
$t^\star\sim|\lambda|^{-1}\log(1/(|\lambda|\mu))$, i.e.\ decreasing in the contraction
rate, so this is a monotone heuristic, not an equality. A cross-check on the
fixed-size corpus: among settling competent runs, solve steps correlate with
$1/|\lambda_{\max}|$ on reachability (Spearman $0.56$, $p{=}10^{-3}$) and weakly,
non-significantly on Sudoku/carry, as expected since $C$ and $\mu$ also vary per instance
and task.

\paragraph{Remark 1 (trajectory informativeness is regime-conditional).}
In branch (a) the certified freeze time is bounded by a decreasing function of the
margin, so per-trajectory statistics encode instance geometry; in branch (b) the freeze
step is an entry time into a level set of $g$ and carries no such information. This is a
statement about \emph{information}, not \emph{success} (a large margin can surround a
confidently wrong state). Empirically, on the settling subset of the fixed-size corpus
($n{=}132$ trajectory files with both correct and incorrect examples), the correct-minus-%
incorrect gap in settle ratio, path length, and effective dimension is directional and
significant (Wilcoxon $p{=}3.4{\times}10^{-9}$, $4.5{\times}10^{-7}$,
$3.9{\times}10^{-11}$); on the marginal subset the gaps vanish (medians $\approx0$); and
on Huginn (marginal) the freeze step does not separate correct from incorrect within a
difficulty rung ($p{\approx}0.9$). This is the mechanism behind the per-instance
confidence read-out of the main text (Sec.~\ref{supp:confidence}); the two use different
corpora and tests, so their $p$-values are not comparable.

\paragraph{Proposition 2 (transient/asymptotic decoupling, single linearization).}
For a fixed step Jacobian $J$ with top right singular vector $v$,
\begin{equation}
\lVert Jv\rVert=\sigma_{\max}(J),\qquad \rho(J)<1\ \Rightarrow\ \lVert J^t\rVert\to0
\end{equation}
give the one-step amplification and, by Gelfand's formula, the asymptotic decay. Thus for a non-normal $J$
a large transient gain $\sigma_{\max}$ coexists with asymptotic contraction; clamping
$\sigma_{\max}$ removes the transient that carries the computation, whereas constraining
$\rho$ leaves it intact. This is elementary and concerns one linearization step; the
thin-cone geometry across the trajectory (main text Sec.~6) is measured, not implied by it.

\section{Analysis populations (run-flow)}
\label{supp:runflow}
Different analyses use different corpora; Table~\ref{tab:runflow} defines each population
and the unit of analysis, so counts that differ across the paper (e.g.\ 17 settling runs
in the depth-safety analysis vs.\ larger settling counts elsewhere) are traceable to
distinct campaigns rather than inconsistency. One run $=$ one trained checkpoint
(task$\times$configuration$\times$seed); tiers, depths, and stored trajectory files are
repeated observations within a run, not independent replications.

\begin{table}[h]
\centering\small
\setlength{\tabcolsep}{4pt}
\begin{tabular}{p{0.30\columnwidth}p{0.16\columnwidth}p{0.42\columnwidth}}
\toprule
\textbf{Corpus} & \textbf{Runs} & \textbf{Used for} \\
\midrule
Original ladders (LOO $+$ architectures) & 304 total; 164 competent (17 settle / 129 marginal / 18 drift) & Depth-safety separation (main Sec.~5) and Fig.~\ref{fig:regime}; anisotropy AUC; non-drift extrapolator count \\
\addlinespace
Fixed-size ladders (primary) & 246 & Table~1, depth curves (main Fig.~2), headline conversion numbers \\
\addlinespace
Trajectory topology (original) & 325 files & $\sigma_{\mathrm{eff}}\!\leftrightarrow\!$settling ($r{=}0.75$); per-instance confidence \\
\addlinespace
Trajectory topology (fixed-size) & 1{,}206 files & Rank replication (Spearman $\rho{=}0.69$); Betti-1 shares \\
\bottomrule
\end{tabular}
\caption{Population and unit of analysis for each corpus. Counts differ by design; the
depth-safety statistic and the regime figure below both use the original-ladder competent
corpus (164 runs).}
\label{tab:runflow}
\end{table}

\subsection{Regime-to-outcome aggregate view}
\label{supp:regime}
\begin{figure}[h]
\centering
\includegraphics[width=\columnwidth]{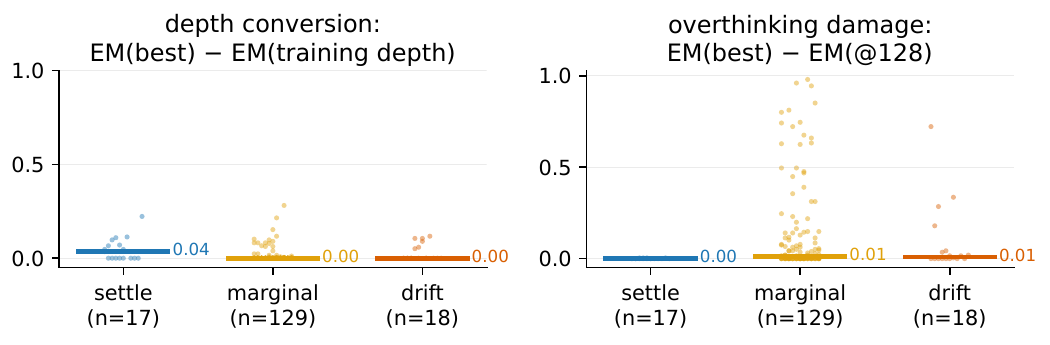}
\caption{Regime $\to$ outcome on the 164 competent original-ladder runs (dot $=$ run,
bar $=$ median; $n$ per regime in axis labels). Left: depth conversion, EM(best)$-$EM at
training depth on the first unseen tier. Right: overthinking damage, worst over tiers of
EM(best)$-$EM@128. Settling runs show no overthinking tail; marginal and drifting runs
do.}
\label{fig:regime}
\end{figure}

Figure~\ref{fig:regime} aggregates the 164 competent original-ladder runs (the same
population as the main-text depth-safety statistic) by measured regime. Settling runs
carry the higher median depth conversion, and overthinking damage concentrates in the
non-settling regimes; conversion is task-dependent and, pooled, not cluster-significant
(main text), so the figure is descriptive of this corpus, not a pooled significance
claim.

\section{The Sudoku ablation landscape in two instruments}
\label{supp:spectra}
\begin{figure}[h]
\centering
\includegraphics[width=\columnwidth]{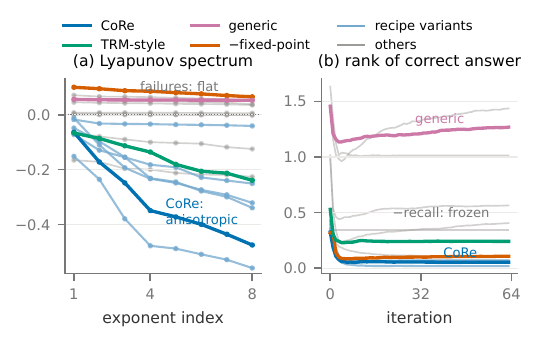}
\caption{The Sudoku ablation landscape (all arms overlaid; colored $=$ narrative-central
operators, light blue $=$ recipe variants, gray $=$ others). \textbf{(a)} Lyapunov
spectra: learners are \emph{anisotropic}; every failure family is flat. \textbf{(b)}
Logit-lens rank of the correct answer (hardest training tier): \textsc{CoRe}
crystallizes to ${\approx}0$, generic bounces, the identity ablation is frozen. Depth
curves for these arms are in the main text; the training-time formation view is in
Sec.~\ref{supp:formation}.}
\label{fig:unified}
\end{figure}

Figure~\ref{fig:unified} overlays every Sudoku arm in the two instruments behind the
anisotropy and crystallization claims of the main text (Secs.~5--6).

\section{Trajectory-topology hypotheses: full statements}
\label{supp:hypotheses}

A topology-first research program \citep{barannikov2022rtd} predicts that reasoning
trajectories fall into settle / loop / drift classes (H1), that winding of the trajectory
tracks reasoning depth (H2), and that a purely contracting map cannot implement
counting-like computation (H3). Our corpus gives verdicts:

\textbf{H1: two of three regimes observed; the third is absent.} Settle and drift are
ubiquitous and sharply separable (main text Sec.~5), and the regime axis is causal via
the fixed-point switch; loops are not observed anywhere, with median total H1
persistence $0.000$ in every regime and Koopman/DMD dominant frequencies are
${\approx}0$ across the corpus.

\textbf{H2: not supported in this task class.} The 99th percentile of $|$winding$|$ is
$0.71$ of a turn, extrapolators included: propagation solvers move radially toward their
fixed point, not rotationally (task classes with cyclic state may differ). The same holds
at scale: on Huginn-3.5B, $|$winding$|{>}1$ occurs for only $0.08\%$ of the $2{,}585$
measured token trajectories.

\textbf{H3: supported, with a sharpened mechanism.} The successful operator is
\emph{not} globally contracting: it contracts on average ($\sigma_{\mathrm{eff}}{<}1$)
while strongly expanding along a thin subspace ($\sigma_{\max}{\approx}4$--$5$), which is
where the computation lives (main text Sec.~6). A regularizer sweep that drives the
\emph{typical} (random-direction) Jacobian gain toward a target $\tau$ (the
Jacobian regularizer of \citet{bai2021stabilizing}, repurposed here as a probe of the
learned geometry rather than a stabilizer) confirms this geometry from both sides. Pinning the typical gain at its natural, decaying level
($\tau{=}0.5$) is harmless: training and extrapolation intact ($0.26$--$0.36$), the
thin amplifying cone untouched ($\sigma_{\max}$ still ${\approx}4$), while forcing
typical directions to \emph{expand} ($\tau{\geq}2$) destabilizes the operator
($\sigma_{\max}$ blows up to $14$--$33$) and destroys training itself. Intermediate
targets already degrade seed-wise: at $\tau{=}1$ one seed in three collapses to
$0.004$ while the others extrapolate normally ($0.37$). Random directions
must decay; only the thin cone may amplify. Regularizing the spectral \emph{radius}
$\rho$ (constraining asymptotics while leaving the transient intact) is likewise
the control that helps at scale \citep{stars2026}.

\section{Non-normality account and estimator scope: full statements}
\label{supp:nonnormal}

Why is $\sigma_{\max}$ a correlate but not a lever? Successful operators have
$\sigma_{\max}/\rho\approx4$--$5.4$ (Henrici index $0.08$--$0.10$ on Sudoku): they are
strongly non-normal, so their top singular value measures a \emph{transient, feedforward}
amplification each step applies to fresh input, not an asymptotic growth rate (which is
$\rho{<}1$). Clamping $\sigma_{\max}$ therefore removes the amplification that carries
the computation; regularizing $\rho$ \citep{stars2026} constrains only the asymptotics,
which is exactly the part that should be controlled. Two refinements from direct measurement:
first, random-probe transient gain stays ${\le}1$ for the recipe even though
$\sigma_{\max}{\approx}5$, so the amplified subspace is thin: random
perturbations decay monotonically, and computation lives in a low-dimensional amplifying
cone inside a globally contracting map. (Worst-case gain requires directed probes;
random probes measure the typical case, a methodological point we make explicit.)
Second, non-normality without convergence control does not help: Neural GPU reaches
$\sigma_{\max}/\rho{=}17.7$, drifts, and scores zero. The recipe's role is to put strong
non-normality \emph{inside} a settling envelope.

\subsection*{Scope of the estimates}
Beyond the instruments named in the main text, the per-run inventory includes
trajectory length, Koopman/DMD dominant frequencies, a Kolmogorov--Sinai entropy proxy,
and gradient statistics at the five training checkpoints; trajectories are stored PCA-reduced to 64 components (metrics use full
$d{=}256$ latents).
We state what each quantity is, so it is read for no more than it measures. The Lyapunov
``spectrum'' is the top eight finite-time exponents (Benettin QR on held-out inputs), not
the full $d{=}256$ spectrum: ``all exponents negative'' is a statement about those eight,
and a Kaplan--Yorke dimension of eight is censored at that cutoff, not an attractor
dimension. Eigenvalue-based quantities (Henrici index, numerical and pseudospectral
radius) are computed on projected blocks and read as sampled, not exact, normality.
$\sigma_{\mathrm{eff}}{=}\exp(\lambda_{\max})$ is a re-expression of the Lyapunov estimate,
not independent corroboration of it. Topological quantities are computed on
PCA-reduced trajectories (median $98.5\%$ variance retained). None of these change the
regime separations, which are large; they bound the adjectives, not the verdicts.

The regime-label threshold $|\lambda_{\max}|{\le}0.05$ is likewise not load-bearing:
relabeling the corpus at thresholds $0.02$, $0.05$, and $0.10$ orders the median depth
conversion of the three regimes identically. As a corpus-level prediction detail, the
Lyapunov spread's AUC $0.846$ for extrapolation compares with $0.390$ for
$|\lambda_{\max}|$ alone: extrapolators are not the most contracting models.

\section{Full leave-one-out table}
\label{supp:fulltable}
Table~\ref{tab:full} extends the main text's Table~1 with every recipe ablation
(none of the six additional arms switches dynamical mode; they degrade gracefully) and
the SE-RRM baseline (fails the competence gate everywhere it was run). Column
definitions are identical to the main text.

\begin{table*}[t]\centering
\setlength{\tabcolsep}{4.2pt}
\resizebox{\textwidth}{!}{%
\begin{tabular}{l ccc ccc ccc}
\toprule
 & \multicolumn{3}{c}{Sudoku (hole fraction)} & \multicolumn{3}{c}{Reachability (path length)} & \multicolumn{3}{c}{Addition (carry length)}\\
\cmidrule(lr){2-4}\cmidrule(lr){5-7}\cmidrule(lr){8-10}
 & learn & extrapolate & overthink & learn & extrapolate & overthink & learn & extrapolate & overthink\\
\midrule
\textsc{CoRe} (full recipe) & 1.00 & 0.19$\to$\,0.34\,{\scriptsize$\pm$0.04} & 0 & 0.98 & 0.92$\to$\,0.92\,{\scriptsize$\pm$0.10} & 0 & 0.99 & 0.13$\to$\,0.17\,{\scriptsize$\pm$0.15} & 0.05\\
\addlinespace[2.5pt]
~~$-$\,structure & 0.00 & 0.00$\to$\,0.00 & 0 & 0.15\,{\scriptsize$\pm$0.02} & 0.08$\to$\,0.08\,{\scriptsize$\pm$0.03} & 0 & \multicolumn{3}{c}{\textit{= CoRe (no spatial structure on 1-D)}}\\
~~$-$\,extra input injection & 0.29\,{\scriptsize$\pm$0.01} & 0.00$\to$\,0.00 & 0 & 0.23 & 0.17$\to$\,0.17 & 0 & 0.04\,{\scriptsize$\pm$0.02} & 0.00$\to$\,0.00 & 0\\
~~$-$\,fixed-point objective & 1.00 & 0.01$\to$\,0.03\,{\scriptsize$\pm$0.01} & 0.04 & 1.00 & 0.97$\to$\,0.97\,{\scriptsize$\pm$0.02} & \textbf{0.37} & 1.00 & 0.14$\to$\,0.15\,{\scriptsize$\pm$0.02} & \textbf{0.81}\\
~~$-$\,orthogonal init & 1.00 & 0.19$\to$\,0.34\,{\scriptsize$\pm$0.03} & 0 & 1.00 & 0.96$\to$\,0.96\,{\scriptsize$\pm$0.01} & 0.05 & 1.00 & 0.18$\to$\,0.32\,{\scriptsize$\pm$0.28} & 0\\
~~$-$\,adaptive step & 1.00 & 0.11$\to$\,0.24\,{\scriptsize$\pm$0.17} & 0 & 0.69\,{\scriptsize$\pm$0.03} & 0.43$\to$\,0.45\,{\scriptsize$\pm$0.18} & 0.11 & 1.00 & 0.39$\to$\,0.40\,{\scriptsize$\pm$0.06} & 0.01\\
~~$-$\,deep supervision & 1.00 & 0.13$\to$\,0.37\,{\scriptsize$\pm$0.04} & 0 & 0.99 & 0.93$\to$\,0.94\,{\scriptsize$\pm$0.02} & 0.07 & 0.99\,{\scriptsize$\pm$0.01} & 0.10$\to$\,0.27\,{\scriptsize$\pm$0.22} & 0.01\\
~~$-$\,grad clip & 1.00 & 0.10$\to$\,0.31\,{\scriptsize$\pm$0.10} & 0 & 0.57\,{\scriptsize$\pm$0.15} & 0.34$\to$\,0.34\,{\scriptsize$\pm$0.19} & 0.05 & 0.99\,{\scriptsize$\pm$0.01} & 0.08$\to$\,0.12\,{\scriptsize$\pm$0.04} & \textbf{0.62}\\
~~lr $3{\cdot}10^{-4}\!\to\!10^{-3}$ & 1.00 & 0.21$\to$\,\textbf{0.42}\,{\scriptsize$\pm$0.01} & 0 & 1.00 & 0.98$\to$\,\textbf{0.98}\,{\scriptsize$\pm$0.01} & 0 & 0.99\,{\scriptsize$\pm$0.01} & 0.27$\to$\,0.58\,{\scriptsize$\pm$0.37} & 0.03\\
~~answer+latent objective & 1.00 & 0.20$\to$\,0.34\,{\scriptsize$\pm$0.04} & 0 & 1.00 & 0.95$\to$\,0.95\,{\scriptsize$\pm$0.02} & 0.01 & 1.00 & 0.07$\to$\,0.11\,{\scriptsize$\pm$0.04} & 0.03\\
\addlinespace[2.5pt]
generic recurrence & 0.00 & 0.00$\to$\,0.00 & 0 & 0.17\,{\scriptsize$\pm$0.01} & 0.07$\to$\,0.11\,{\scriptsize$\pm$0.02} & 0.13 & 1.00 & 0.11$\to$\,0.12\,{\scriptsize$\pm$0.02} & \textbf{0.84}\\
TRM-style & 0.94\,{\scriptsize$\pm$0.18} & 0.14$\to$\,0.21\,{\scriptsize$\pm$0.11} & 0 & 0.70\,{\scriptsize$\pm$0.17} & 0.55$\to$\,0.56\,{\scriptsize$\pm$0.22} & 0.01 & 1.00 & 0.80$\to$\,0.82\,{\scriptsize$\pm$0.15} & 0.02\\
UT/URM & 0.97\,{\scriptsize$\pm$0.06} & 0.00$\to$\,0.00 & 0.18 & 0.43\,{\scriptsize$\pm$0.09} & 0.28$\to$\,0.28\,{\scriptsize$\pm$0.07} & 0.09 & 0.99\,{\scriptsize$\pm$0.02} & 0.66$\to$\,0.72\,{\scriptsize$\pm$0.21} & \textbf{0.59}\\
FPRM & 0.53\,{\scriptsize$\pm$0.40} & 0.00$\to$\,0.01\,{\scriptsize$\pm$0.01} & 0.04 & 0.31\,{\scriptsize$\pm$0.05} & 0.20$\to$\,0.20\,{\scriptsize$\pm$0.03} & 0.11 & 0.82\,{\scriptsize$\pm$0.26} & 0.27$\to$\,0.28\,{\scriptsize$\pm$0.25} & \textbf{0.80}\\
EqR & 1.00 & 0.01$\to$\,0.01\,{\scriptsize$\pm$0.01} & 0.13 & 0.99\,{\scriptsize$\pm$0.01} & 0.88$\to$\,0.89\,{\scriptsize$\pm$0.03} & \textbf{0.24} & 1.00 & 0.12$\to$\,0.12\,{\scriptsize$\pm$0.03} & \textbf{0.72}\\
DEQ & 1.00 & 0.00$\to$\,0.00 & 0 & 0.26\,{\scriptsize$\pm$0.04} & 0.16$\to$\,0.16\,{\scriptsize$\pm$0.03} & \textbf{0.24} & 0.99\,{\scriptsize$\pm$0.01} & 0.19$\to$\,0.47\,{\scriptsize$\pm$0.19} & 0.07\\
Neural GPU & 0.00 & 0.00$\to$\,0.00 & 0 & 0.76\,{\scriptsize$\pm$0.08} & 0.54$\to$\,0.54\,{\scriptsize$\pm$0.11} & \textbf{0.68} & --- & --- & ---\\
SE-RRM & 0.00 & 0.00$\to$\,0.00 & 0 & --- & --- & --- & --- & --- & ---\\
\bottomrule
\end{tabular}}
\caption{Full leave-one-out and architecture table (mean$\pm$sd; architecture rows
$n{=}10$ seeds, \textsc{CoRe}/$-$fixed-point/TRM-style $n{\approx}20$, other ablations and
SE-RRM $3$); column
definitions as in main text Table~1. \textbf{learn}: in-distribution EM at best depth.
\textbf{extrapolate}: first-unseen-tier EM at training depth $\to$ best over
$h_c{\le}128$. \textbf{overthink}: worst over tiers of best EM $-$ EM@128 (bold
${\geq}0.2$); ``---'' marks arm undefined or not run.}
\label{tab:full}\end{table*}

\section{Huginn-3.5B: measurement details}
\label{supp:layernorm}
\paragraph{R2 within-family detail.} On the last-digit ladder the median freezing step
is $12/12/10/12$ across operand sizes ($1$--$4$ digits): stabilization does not track
difficulty \emph{within} a family either, not only across them (main text Sec.~7).
On the chain ladder (first rung within competence) accuracy decays $0.72/0.40/0.23$
while the freezing step stays put and plateaus by $r{\approx}8$--$16$.
\paragraph{Probe scope.} The behavioral probes are small (304 prompts) and test the
R1--R4 criteria, not model ranking; the base model is prompted zero-shot, so absolute
accuracies are lower bounds. Across the $2{,}585$ per-token trajectories the effective
dimension still tracks task category ($2.23\to2.68$, trivial to multistep), so the
iteration is not inert; the bounded, orbit-like trajectories are consistent with what
the model's authors report \citep{geiping2025huginn}.
\paragraph{The layer-normalization objection, in full.}
Is $\sigma_{\mathrm{eff}}{=}1.000$ not simply what layer normalization enforces in a
pre-norm transformer? Partly, yes, and we lean on it accordingly. Normalization pins
the radial direction of the state, so a Lyapunov estimate near zero is partially
expected \emph{regardless} of what the map computes, and we do not rest the verdict on
it. The verdict rests on trajectory \emph{displacement}, which normalization does not
fix: an operator settling toward a fixed point \emph{on} the norm shell would show
per-step displacement decaying to zero, and none of the $2{,}585$ measured token
trajectories does: the latent keeps traveling at an undiminished rate while the
decoded answer sits frozen (answer freezing at median 3--5 changes; main text Sec.~7). This
is exactly the premise of Proposition~1(b): a frozen answer with non-vanishing
displacement carries no depth-safety guarantee.
Normalization explains why marginality is the \emph{default} for this architecture
family; it does not explain why training never leaves it; our small-scale
constructive experiment (main text Sec.~5) shows a single objective term suffices to leave
it when the objective asks.

\section{Repairing depth-safety at scale: protocol, transfer laws, negative arms}
\label{supp:repair}
All runs in this section were pre-committed before execution (protocol files in the
repository); every result file encodes its full configuration.

\paragraph{The base pathology.} On procedural single-token probes the pretrained
model's accuracy peaks at a shallow, task-dependent recurrence
($r^{*}{\in}[2,24]$; carry ${\approx}8$, digit-copy $2$, index-digit $24$) and then
collapses: carry $0.31\,(r{=}4) \to 0.69\,(r{=}8) \to 0.00\,(r{\ge}32)$, copy dead from
$r{=}4$, letter-repeat $0.56 \to 0.00$. At depth every input freezes onto one
input-independent wrong token, so the failure is a settling event
(Proposition~1b). Consequently answer-agreement halting cannot detect it: stopping when
the decoded answer stops changing recovers $0.56$ on carry against $0.74$ for an oracle
fixed depth, and always-deep inference yields $0.00$; a 30-example calibration of a
fixed depth recovers the peak.

\paragraph{Repair objective.} A rank-16 LoRA adapter on the recurrent core is trained
with the single loss
$\|z(x,r) - z_{\mathrm{base}}(x,t^{*})\| / \|z_{\mathrm{base}}(x,t^{*})\|$
on the answer position, $r{\sim}U[16,128]$, teacher $t^{*}{=}8$, learning rate
$10^{-5}$, $2{,}000$ steps, no task labels; anchors are two synthetic domains
(smallest-digit, first-letter). Held-out results over $10$ seeds: digit-copy
$0.00 \to 1.00{\pm}0.00$ at $r{=}128$ (all seeds input-dependent by prediction
entropy); letter-repeat $0.00 \to 0.51{\pm}0.13$; carry tier~8 plateau
$0.60{\pm}0.04$ (tiers $2/16$: $0.48/0.53$) with largely input-independent answers,
which we count as stabilization, not restored computation. An 11-domain anchor variant
reaches carry $0.68{\pm}0.10$ ($4/9$ seeds input-dependent) and holds copy at
$0.93$ when evaluated at $r{=}256$; the two-domain recipe is verified to $r{=}128$.
A single first-letter anchor gives the best carry plateau we observed
($0.60$--$0.96$ deep; two seeds, one clearly input-dependent).

\paragraph{Transfer follows the anchors' operation, across symbol alphabets.} A single
word-morphology anchor (pluralization, which echoes its stem) restores copying in
digits ($0.98{\pm}0.00$) and letters ($0.89{\pm}0.08$), input-dependent in $6/6$
seeds; carrier cross-tests are diagonal (smallest-digit repairs digit-copy $1.00$ but
letter-repeat only $0.06$; first-letter repairs letter-repeat $0.45$ but digit-copy
$0.25$). Non-echo anchors restore nothing: a fact-association anchor with a stronger
base peak than pluralization's ($0.27$ vs.\ $0.17$) repairs $0.00$, and random-string
corpora likewise, so the operation, not anchor strength, carries the transfer.

\paragraph{Two failure laws, measurable in advance.} (i)~Anchor validity: domains the
base itself cannot perform poison the mix. Adding four counting domains (base
near-chance) to the two-anchor recipe drops copy from $1.00$ to $0.50{\pm}0.15$; an
association domain the base scores $0.00$ on drives the pluralization anchor's copy
transfer from $0.98$ to $0.00$. A 30-example base scan of each candidate anchor gates
this before training. (ii)~Mirror interference: a near-duplicate template of an anchor
with contrary semantics is damaged by state capture, not semantic capture: with a
smallest-digit anchor, largest-digit falls $0.58 \to 0.06$--$0.09$ deep ($90\%$ of its
deep answers are not digits at all), and the damage tracks the twin's share of the
training mix; last-letter's peak falls $0.13 \to 0.00$ under a first-letter anchor
while a non-twin anchor keeps it at $0.14$. With no twins among the anchors, no
evaluation task is damaged.

\paragraph{What never appears: conversion.} No treated model exceeds the base's peak
competence at any depth, in any arm. The depth-hard pointer-chase task (iterated
single-cycle permutation, $S{=}10$) is unlearnable in-distribution under LoRA rank
$16$--$128$ and full-core tuning at learning rates $3{\cdot}10^{-6}$--$10^{-4}$
(accuracy at chance, and never increasing with $r$). The fixed-point loss itself does
not transfer to this scale: trained alongside the task it merely freezes an
already-frozen loop, and trained alone it collapses the model to a constant answer at
every dose we tried, while as little as $50$ gentle steps already flatten the base's
depth profile. Where a trained arm does show settling together with high flat accuracy
(e.g., a task-trained control at $0.84$ on carry with displacement $10^{-5}$), the
answers are identical across tiers and depths, an input-independent constant rather
than competence. Behavioral distillation from the base's peak depth (a per-domain
teacher) repairs a seven-task battery in-mix ($5/5$ pre-committed criteria; the
per-domain teacher matters for copy $0.65$ vs.\ $0.00$ and largest-digit $0.62$ vs.\
$0.41$ against a uniform teacher) but never holds the deep tail on held-out domains;
only the frozen latent anchor transfers.

\paragraph{Seed variance and a label-free selector.} Depth-contract training inherits
the model's stochastic latent initialization; where seeds vary (the 11-domain variant),
degenerate seeds are detectable without labels by prediction entropy at depth
(top-class fraction ${\ge}0.85$): the selector keeps $9/10$ seeds on copy (mean $0.90$,
dropping exactly the dead seed). The two-anchor recipe required no selection
($10/10$).

\section{Weak-form R4: the full natural experiment}
\label{supp:weakr4}
The size-ladder experiment separating the two forms of answer convergence (main text
Sec.~5): on addition-by-\emph{length} the generic recurrence length-generalizes
($0.83\!\to\!0.86$; a size-axis result, not an R1 claim) with an answer frozen within
${\sim}5$ steps, \emph{while its latent never settles}
($\sigma_{\mathrm{eff}}{=}1.02$, still moving at the last step): the carry front
travels through the state in directions the readout ignores (weak-form R4). On prefix
sums the \emph{same} operator statistics ($\sigma_{\mathrm{eff}}{=}1.02$,
$\lambda_{\max}{=}{+}0.02$, spectrally indistinguishable from the addition twin)
fail R4: the generic net learns the training tier better than the recipe ($1.00$ vs.\
$0.23$ at depth 16) and is then destroyed by test-time depth ($1.00\!\to\!0.65$ on the
same tier, $0.56\!\to\!0.04$ on the next), while the recipe's weaker solution is
depth-stable ($0.23\!\to\!0.26$).

\paragraph{Strict R2 on a deterministic oracle.} The TRM-style model's halting on
addition-carry tracks the oracle's carry length nearly one-to-one: $16$-round carry
chains stop at $14.5$--$15.1$ solve steps (main text Sec.~5).

\section{The fixed-point objective's gate does not cheat}
\label{supp:gate}
The fixed-point penalty is scaled by the learned update gate $a_T$, so the gate could
in principle satisfy the penalty by closing ($a_T\to0$) without the map learning a
fixed point. Trained recipes do not do this: they keep a small but nonzero terminal
update and difficulty-adaptive solve steps, unlike the flat-at-$2.0$ solve steps of the
identity mode (Sec.~\ref{supp:modes}).

\section{Objective interventions: dose, matched control, beyond-horizon}
\label{supp:interventions}
These experiments support the causal analysis of main text Sec.~5.4; protocols were
committed before the runs (Sec.~\ref{sec:repro}).

\paragraph{Answer-stabilization control (isolating the latent attractor).} We train the
recipe with the residual penalty redirected from the latent to the \emph{decoded answer}
(penalizing $\lVert \mathrm{readout}(z_T)-\mathrm{readout}(z_{T-1})\rVert$, leaving the
latent free). This freezes the output without asking the latent to converge. It reproduces
neither payoff: on Sudoku it fits its easy training tiers (hardest train tier only
$0.18$--$0.30$) and extrapolates to $0.00$ on all three seeds, with the latent never
settling ($\lambda_{\max}{=}{+}0.03$--$0.07$); on reachability it matches the recipe's
unseen-tier peak ($0.91$--$0.95$) but is then destroyed by depth (worst-tier overthinking
damage $0.99$--$1.00$, versus $0.00$ for the latent-settling recipe).
The payoff therefore requires the latent fixed point the objective creates, not answer
freezing.

\paragraph{Dose-response of the objective weight.} Sweeping the objective coefficient
(three seeds each; the headline weight is $0.05$):
\emph{carry propagation, generic recurrence}: at weight $0$ the operator drifts
($\lambda_{\max}{=}{+}0.04$), with depth damage $0.87$--$0.93$ and conversion only
$0.09$--$0.14$; at $0.005$ the exponent crosses into settling and the damage vanishes
($0$ of $3$ unsafe) while conversion is still small; at $0.02$ conversion appears
($0.60$--$0.97$); at $0.05$--$0.2$ it saturates (${\approx}0.88$, matching the headline; $3$-seed dose points). Adding the objective
thus fixes both the damage and the missing conversion; depth-safety switches on at the
settling crossing, and conversion then grows with weight.

\emph{Sudoku, recipe}: weights $0.01$ and $0.05$ are comparable
(extrapolation $0.32$--$0.41$), weight $0.2$ suppresses it ($0.05$--$0.16$), and weight
$1.0$ over-contracts into a deep settle ($\lambda_{\max}{=}{-}0.26$--$-0.36$) with perfect
training accuracy but zero extrapolation. Settling is necessary but not sufficient: the
useful target is the slowest rate that still settles.

\emph{EqR, native weight}: the same over-dose boundary appears architecturally. EqR
hard-wires this penalty at weight $0.1$, yet on these ladders its residual never resolves
(the training loss plateaus at the penalty term while in-distribution EM is perfect), and
the model overthinks instead of converting (Table~\ref{tab:full}). With the native term
removed, the unchanged backbone converges and converts (carry $0.68{\to}1.00{\pm}0.01$,
Sudoku $0.14{\to}0.34{\pm}0.04$; $n{=}10$ each), the trained operator settling on its own
on carry. The objective pays only where the penalty can actually be driven down; a
hard-wired weight over a residual that stays large buys neither safety nor conversion.

\paragraph{Constructive transfer across architectures.} Re-seeded to $n{=}10$ per cell,
adding only the fixed-point term (weight $0.05$) to the two marginal architectures flips
the measured regime seed-by-seed: on carry and reachability every UT/URM and FPRM control
run has $\lambda_{\max}{>}0$ and every $+$fp run $\lambda_{\max}{<}0$ ($40/40$ runs), and
overthinking is repaired wherever there is competence to protect (carry damage
$0.59\!\to\!0.17$ for UT/URM, $0.80\!\to\!0.14$ for FPRM). Conversion does not come along
automatically: it rises on FPRM ($0.28\!\to\!0.49$) but falls on UT/URM
($0.72\!\to\!0.54$), and on reachability the term costs trainability itself (learn
$0.43\!\to\!0.18$ and $0.31\!\to\!0.13$, below the competence gate). DEQ marks the other
boundary: its control already settles on carry ($\lambda_{\max}{<}0$, $10/10$), and adding
the term stops training from fitting at all ($10/10$ seeds). Under the intervention the
regime and depth-safety move together on foreign architectures, so the arm's identity does
not carry them; conversion stays task- and architecture-dependent, as in the main text
(settling is necessary, not sufficient).

\paragraph{Beyond-horizon conversion on a deterministic oracle.} To test conversion
strictly past the training horizon we retrain at $T_{\mathrm{train}}{=}8$ and evaluate on
oracle depths that exceed it (carry length up to $28$, i.e.\ $3.5\times$ the horizon).
Across $25$ seeds the at-horizon tier (carry-$8$) converts on $17$ (mean EM $0.59$), but
the strictly-beyond-horizon tiers do not convert: the $16$-round tier exceeds EM $0.5$ on
$1$ of $25$ seeds (mean $0.06$) and the $28$-round tier on none (mean $0.02$). An initial
five-seed run had a single seed convert the deep tiers ($0.88$/$0.46$); this did not
replicate at scale, so we make no strictly-beyond-horizon conversion claim on carry. What
is robust strictly past the horizon is depth-safety: the settling recipe holds a flat
plateau where its no-fixed-point twin peaks higher and then collapses (the safety contrast
is sharper than the conversion contrast, as at $T_{\mathrm{train}}{=}16$).

\section{Failure-mode portraits}
\label{supp:modes}
Full quantitative portraits of the four failure modes (main text Sec.~5):

\emph{The identity mode} (no-recall): the loop converges instantly to ``do nothing''
($\sigma_{\max}{=}1.0000$; solve steps $2.0$ flat, the only arm with no
difficulty-adaptivity at all; rank curve frozen). R2--R4 all fail in the most degenerate
way; accuracy $0.00$.

\emph{The marginal mode} (generic recurrence, UT/URM, FPRM, SE-RRM): the trained operator
is close to normal and norm-preserving (Henrici departure $0.000$,
$\sigma_{\max}{=}\rho{=}1.000$), with a near-zero-to-weakly-positive
$\lambda_{\max}$ sitting at or just above the marginal band (values around
$+0.05$--$0.07$ straddle the marginal/drift threshold: these operators are the least
settled non-drift family, not cleanly interior to marginal), saturated Kaplan--Yorke
dimension, and positive entropy proxy. Latents wander a high-dimensional filament
(trajectory effective dimension ${\approx}1.1$, path length ${\sim}15$k); rank curves
stay high and non-monotone. On the original ladders these models fit easy tiers (UT/URM
reaches $0.56$ training-tier EM on those tasks; on the fixed-size ladders of Table~1 the
same family reaches $0.97$; the two figures are different corpora, Sec.~\ref{supp:runflow})
but R1 fails: their depth curves are flat or declining.

\emph{The drift mode} (no fixed-point objective): non-normality survives
($\sigma_{\max}{=}5.6$) but $\lambda_{\max}{=}+0.10$
($\sigma_{\mathrm{eff}}$ $0.94\!\to\!1.11$ relative to the recipe); the trajectory is ten times longer
than the recipe's, entropy is the highest measured, and extra test-time steps actively hurt
(R4 fails with consequences for R1: $0.33\to0.04$).

\emph{The settle-but-wrong mode} (no-structure; TRM on ARC): the dynamics are right,
as no-structure contracts hardest of all arms; TRM on ARC settles with
$\lambda_{\max}{=}-0.21$ and $\sigma_{\max}{=}30$--$60$), but the fixed point does not
encode the solution (EM $0.00$ and $0.013$ respectively). Convergence control cannot
substitute for an inductive bias aligned with the algorithm (main text Sec.~4).
Spectrally these runs settle \emph{isotropically} (Lyapunov spread ${\le}0.08$), whereas
true algorithm-learners settle \emph{anisotropically} (spread $0.2$--$0.4$); the
instrument that separates this mode from success is developed in
main text Sec.~6.

\section{Practitioner's dashboard}
\label{supp:dashboard}
Table~\ref{tab:dashboard}: what to monitor while training a recursive model, what each
pathology looks like, and which recipe component treats it (main text Sec.~6).

\begin{table}[h]
\centering
\small
\begin{tabular}{p{0.30\columnwidth}p{0.30\columnwidth}p{0.26\columnwidth}}
\toprule
\textbf{Signal} & \textbf{Diagnosis} & \textbf{Fix} \\
\midrule
$\sigma_{\max}\to1.0000$, solve steps flat & identity fixed point (``do nothing'') & extra input injection (recall) \\
\addlinespace
Henrici $\to0$, $\sigma_{\max}{=}\rho{=}1$, KY saturated & normal marginal wander (coverage learner) & structured step; fixed-point objective \\
\addlinespace
$\lambda_{\max}{>}0$, trajectory length exploding & drift; depth will hurt at test time & fixed-point objective \\
\addlinespace
settles but EM $=0$; flat Lyapunov spectrum ($\lambda_1{-}\lambda_8{<}0.1$) & settled isotropically: converged to a non-algorithm & inductive bias (structure) / less augmentation, not more dynamics control \\
\addlinespace
all eight estimated leading exponents $<0$, rank curve monotone $\downarrow$, EM rises with depth & learned recursion & none needed \\
\addlinespace
seed-to-seed EM variance $\gg$ & missing step-size adaptation / clipping & adaptive updates; clip \\
\bottomrule
\end{tabular}
\caption{Signal $\to$ diagnosis $\to$ fix: what to monitor while training a recursive
model, the pathology each pattern indicates, and the recipe component that treats it.}
\label{tab:dashboard}
\end{table}

\section{Corpus regularities: full statements}
\label{supp:corpus}
\paragraph{One settle axis, measured two independent ways.} Live spectral measurements
and post-hoc trajectory geometry agree run-by-run: $\sigma_{\mathrm{eff}}$ correlates
with trajectory settling at $r{=}0.75$ ($n{=}325$, $p{=}4{\times}10^{-60}$) on the
original campaign and replicates on the fixed-size-ladder corpus as a rank correlation
(Spearman $\rho{=}0.69$, $n{=}1{,}206$; rank-based because a few deeply-settled runs
distort a Pearson estimate).
\paragraph{Regime statistics.} Across the 304 original-ladder runs the 19
extrapolating runs on constraint-propagation tasks separate from the rest on both
dynamical axes: $\lambda_{\max}$ (median $-0.047$ vs.\ $+0.009$,
$p{=}7{\times}10^{-6}$) and Kaplan--Yorke dimension ($0$ vs.\ $8$,
$p{=}2.6{\times}10^{-5}$).
\paragraph{Topology.} A settled cloud is compact, a drifting one a filament: the
near-zero-persistence Betti-1 bar count is positive for $92\%$ of settle-regime
trajectory files versus $24\%$ marginal and $2\%$ drift. Within the same run, correct
examples live on lower-dimensional, more compact trajectories (Sec.~\ref{supp:confidence}).
\paragraph{Formation caveats.} Mid-training hidden-state stable rank predicts final
extrapolation at AUC $0.774$, but arm identity partially confounds, so it is a monitoring
candidate, not a law; and we do not claim the canonical delayed test-accuracy jump of
grokking (the checkpoint grid is too coarse). See also Sec.~\ref{supp:formation}.
\paragraph{Anisotropy: instrument-level exception.} The no-adastep arm extrapolates
with a compressed Lyapunov spectrum, since adaptive update scaling may deform the
measurement, so the anisotropy claim is read as a corpus regularity, not a per-run
guarantee.
\paragraph{History in the spectrum.} Isometrically initialized models end with
double-to-triple the weight stable rank of their ablation at indistinguishable accuracy
so persistent spectral differences can encode history, not just competence.

\section{Training-time formation}
\label{supp:formation}
\begin{figure}[h]
\centering
\includegraphics[width=0.8\columnwidth]{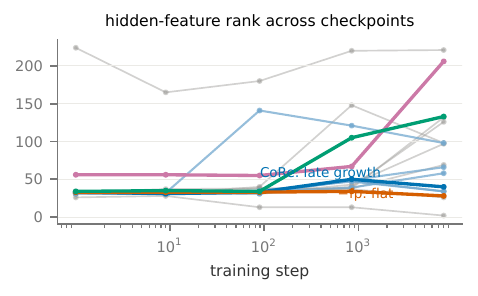}
\caption{Hidden-feature rank across training checkpoints (Sudoku; colors as in the main
text).}
\label{fig:formation}
\end{figure}

Figure~\ref{fig:formation}: hidden-feature rank across the five training checkpoints.
Arms that will extrapolate grow rank late; the no-fixed-point ablation finds its shortcut
early and stays flat (main text Sec.~6). This is the inference-map face of delayed
generalization: as in grokking \citep{power2022grokking}, the generalizing solution
forms long after the loss is low, visible in hidden progress measures before behavior
\citep{nanda2023progress,barak2022hidden}, competing with a cheaper memorizing solution
\citep{varma2023circuit}, the coverage solutions of the main text. The quantities a
grokking analysis would monitor on the training map (rank, spectral structure) are
exactly the ones that diagnose the trained inference map. Within the same runs, correct
examples additionally live on lower-dimensional, more compact trajectories
($p{=}3.5{\times}10^{-10}$ and $p{=}2.3{\times}10^{-6}$ within-run;
Sec.~\ref{supp:confidence}).

\section{Per-instance confidence read-out}
\label{supp:confidence}
The per-instance predictor of the main text (Sec.~6) is a logistic read-out of five
per-trajectory quantities (settling ratio, effective dimension, path length, winding,
Koopman magnitude) with grouped cross-validation over runs (no run leaks across folds).
Cross-validated AUC: $0.80$ (Sudoku), $0.91$ (reachability), $0.98$ (size-Sudoku);
correct examples settle more, on shorter, lower-dimensional paths
($p{=}3.5{\times}10^{-10}$ and $p{=}2.3{\times}10^{-6}$ within-run).

\section{Data generation}
\label{supp:datagen}
Every generator is deterministic given a seed, verifies its oracle-round invariant on each
instance at generation time, and materializes splits once that are shared byte-for-byte
across arms. We describe the three primary fixed-size ladders; the remaining generators
follow the same contract (fixed length, vocabulary, and grid across a ladder, with only the
oracle-round count growing) and are included in the code supplement.

\paragraph{Sudoku (hole fraction).} For each instance we draw a complete valid
$9{\times}9$ solution by randomized backtracking, then blank a fraction $h$ of the $81$
cells chosen uniformly at random ($\lfloor 81h\rfloor$ cells set to the blank token). The
givens are exactly the un-blanked solution cells, so every puzzle is consistent with its
solution by construction. Difficulty is the hole fraction $h$ (training $0.2/0.35/0.5$,
evaluation $0.2$--$0.7$). Uniqueness is not enforced at generation: we compute the
uniquely-solvable fraction per tier ($1.00/0.91/0.50/0.06/0.00$) with an exhaustive solver
and score predictions with the validity oracle, reporting reference match alongside. The
deterministic-oracle ladders below have no such ambiguity.

\paragraph{Reachability (BFS depth).} Undirected graphs on $N{=}16$ nodes with a fixed
reachable set of $R{=}13$. From a source we build $d$ non-empty layers, place at least one
edge from each node to the previous layer, and add adjacent-layer edges with probability
$0.25$, so the BFS depth of every reachable node equals its layer index and the farthest is
exactly $d$. The remaining $N-R{=}3$ nodes form a separate component (internal edge
probability $0.4$), so the reachable fraction $R/N$ is constant across tiers and carries no
base-rate signal. Node identities are permuted at random, and each instance is verified
(BFS depth $=d$) before it is kept. Difficulty is the exact BFS depth $d$ (training $2,3$;
evaluation $2,3,5,8,12$). The input is the flattened adjacency with the source marked on
its diagonal; the target is the per-node reachable bit.

\paragraph{Addition (carry-run length).} Two $32$-bit operands in an interleaved
least-significant-bit-first encoding ($L{=}64$; sum bit $i$ at position $2i$, odd positions
padding), which keeps the carry chain strictly local. Difficulty is the longest run $c$ of
consecutive live carries: we plant one generate-then-propagate cascade of length exactly
$c$ and absorb any accidental longer run, verifying $c$ by carry simulation. The
oracle-round count equals $c$ (a local ripple needs $c$ sequential steps) and is
recomputable from the input. As pre-committed, the fraction of propagate columns grows with
$c$, as the hole fraction does along the Sudoku ladder.

\section{Reproducibility details}
\label{sec:repro}
Every configuration trains at $16$ recurrent steps with deep supervision ($4$ supervised
unrolls), AdamW at learning rate $3{\times}10^{-4}$, gradient clipping $0.5$, batch $64$,
hidden width $d{=}256$, $800$ unique unaugmented examples per difficulty tier, and three
seeds; evaluation uses $256$ fixed examples per tier and sweeps test-time depth
$h_c\in\{3,16,32,64,128\}$. Data files are materialized once and shared byte-for-byte
across every arm and architecture. Table~\ref{tab:repro} lists, per task, the difficulty
parameter, the training and evaluation tiers (as parameter values), and the training-step
budget. Eight runs fell back to a halved batch after an out-of-memory event and continued
from there; the actual batch size is logged per run, so no run silently changes budget.
By ``pre-committed'' we mean that each campaign's protocol, arm list, and headline
choice were committed to version control before the corresponding run and fixed there
rather than chosen after inspecting results; we say ``pre-committed'' rather than
``pre-registered'' because the commits are not timestamped by a third party.

\paragraph{Availability.} An anonymized code supplement accompanies the submission:
generator code and materialized splits with checksums, exact seeds and per-arm
configurations, the measurement and estimator code with the numerical settings of
Sec.~\ref{supp:nonnormal}, architecture implementations, per-run and per-example logs,
and scripts that regenerate every table and figure. It also lists the eight runs that
fell back to a halved batch after an out-of-memory event, with their effective budgets.

\begin{table}[h]
\centering\small
\setlength{\tabcolsep}{4pt}
\resizebox{\columnwidth}{!}{%
\begin{tabular}{llccc}
\toprule
task & difficulty & train & eval & steps\\
\midrule
\multicolumn{5}{l}{\emph{Fixed-size ladders (depth axis; primary)}}\\
sudoku & hole fraction & 0.2,0.35,0.5 & 0.2--0.7 & 8k \\
reach\_pathlen & BFS depth & 2,3 & 2,3,5,8,12 & 4k \\
addition\_carry & carry-run len & 2,4 & 2,4,8,16,28 & 4k \\
maze\_pathlen & path length & 6,12 & 6,12,24,48,80 & 6k \\
sort\_disorder & displacement & 2,4 & 2,4,8,16,24 & 4k \\
maze\_pathlen\_m & path length (matched) & 8,12 & 8,12,16,24,32 & 6k \\
sort\_disorder\_m & displacement (matched) & 2,4 & 2,4,6,8,12 & 4k \\
\addlinespace
\multicolumn{5}{l}{\emph{Size/length ladders (scope; main text Sec.~3)}}\\
sudoku\_size & board size & 9 & 4,9,16,25 & 6k \\
reachability & \#nodes & 8,12 & 8,12,16,24 & 4k \\
maze & grid size & 11,15 & 11,15,21,31 & 6k \\
addition & operand bits & 8,16 & 8--128 & 4k \\
prefix & bit length & 32,64 & 32--512 & 4k \\
parity & bit length & 32,64 & 32--512 & 4k \\
sort & length & 8,16 & 8,16,32,64 & 4k \\
\bottomrule
\end{tabular}}
\caption{Per-task difficulty parameter, training/evaluation tiers (parameter values), and
training-step budget. All other hyperparameters are shared (text above).}
\label{tab:repro}
\end{table}

\end{document}